\documentclass[runningheads,a4paper]{llncs}
\usepackage{amsmath} 

\usepackage{amsfonts}

\title{A Hybrid Computational Intelligence Framework for scRNA-seq Imputation: Integrating scRecover and Random Forests}

\author{Ali Anaissi$^{1,2*}$, Deshao Liu$^{3,4,5}$, Yuanzhe Jia$^{2}$ , Weidong Huang$^{1}$, Widad Alyassine$^{2}$  and Junaid Akram$^{2}$}

\institute{University of Technology Sydney, Australia\\ \and
University of Sydney, Australia \\ \and
Asia Pacific International College (APIC), Parramatta, NSW, Australia\\ \and
Lincoln Institute of Higher Education (LIHE), Sydney, Australia\\ \and
The Institute of International Studies (TIIS), Sydney, Australia  \\ 
\email{ali.anaissi@uts.edu.au, Deshao.Liu@ieee.org, yjia5612@uni.sydney.edu.au, weidong.huang@uts.edu.au, widad.yassien@gmail.com, junaid.akram@sydney.edu.au}}

\begin{document}

\maketitle
\thispagestyle{empty}
\pagestyle{empty}

\begin{abstract}
Single-cell RNA sequencing (scRNA-seq) enables transcriptomic profiling at cellular resolution but suffers from pervasive dropout events that obscure biological signals. We present \textbf{SCR-MF}, a modular two-stage workflow that combines principled dropout detection using \textit{scRecover} with robust non-parametric imputation via \textit{missForest}. Across public and simulated datasets, SCR-MF achieves robust and interpretable performance comparable to or exceeding existing imputation methods in most cases, while preserving biological fidelity and transparency. Runtime analysis demonstrates that SCR-MF provides a competitive balance between accuracy and computational efficiency, making it suitable for mid-scale single-cell datasets.

\end{abstract}

\section{Introduction}
Bulk RNA sequencing aggregates transcript counts over cells, obscuring cell specific programs and temporal dynamics \cite{peng2019scrabble}. Single‑cell RNA sequencing (scRNA-seq) overcomes this limitation by quantifying messenger RNA from individual cells, thereby exposing cellular heterogeneity and dynamic transcriptional changes. 
However, scRNA‑seq count matrices are extremely sparse, with many zeros even for genes that are truly expressed. 
These dropouts confound downstream analyses; thus, a central challenge is to distinguish technical zeros from genuine biological absence so that imputation can recover missing signal without fabricating expression. 

A broad spectrum of imputation strategies has emerged, often leveraging highly variable genes, neighborhood graphs, or low‑dimensional embeddings to borrow strength across similar cells \cite{hou2020systematic}. Yet many approaches are only weakly integrated with downstream tasks and can inflate false positives in differential expression or marker discovery.
These limitations motivate methods that (i) recover expression accurately while (ii) preserving true biological zeros and (iii) scaling to large cell numbers.

In this study, we revisit the imputation pipeline through a two‑stage design: first identifying likely dropouts and then imputing their values. Concretely, we benchmark representative approaches, including established methods, scImpute \cite{li2018accurate}, scRecover \cite{miao2019screcover}, VIPER \cite{chen2018viper}, and MAGIC \cite{dijk2017magic}, as well as composite variants. 

Against this backdrop, we introduce and evaluate our proposed procedure, SCR-MF, which couples scRecover's dropout detection with a random‑forest‑based imputation step. Our contributions are threefold. 
(i) We present a practical, modular imputation framework that explicitly separates dropout detection from value recovery. 
(ii) We conduct a more comprehensive evaluation than prior appraisals by comparing a larger panel of imputation variants and examining their impact on downstream analyses. 
(iii) We assess performance with standard clustering metrics and information‑theoretic similarity, enabling consistent comparisons across datasets.

The paper is organized as follows. 
Section \ref{sec:relatedwork} reviews related work and situates our contribution. 
Section \ref{sec:method} details the methodology and model design. 
Section \ref{sec:experiment} describes datasets, training procedures, evaluation metrics, and exploratory analyses used to validate robustness and applicability. 
Section \ref{sec:conclusion} concludes and outlines future directions.

\section{Related Work}
\label{sec:relatedwork}

\begin{table}
\caption{Notations used in this paper.}
\label{tab:notations}
\centering
\begin{tabular}{ p{3cm} p{8cm} }
\hline
Abbreviation & Description \\
\hline

ALRA & Adaptively-thresholded Low Rank Approximation \\
ARI & Adjusted Rand Index \\
bayNorm & Bayesian Gene Expression Recovery \\ 
DCA & Deep Count Autoencoder \\
DeepImpute & Deep Neural Network-based Imputation \\
DrImpute & Imputation approach for estimating dropout events in scRNA-seq data \\
GEO & Gene Expression Omnibu \\
KNN-smoothing & K Nearest Neighbor Smoothing \\
MAGIC & Markov Affinity-based Graph Imputation of Cells \\
mcImpute & Matrix Completion Based Imputation \\
mRNA & Messenger RNA\\
NCBI & National Center for Biotechnology Information \\
NMI & Normalized Mutual information \\
OOB & Out-of-bag \\
PCA & Principal Component Analysis \\
RF & Random Forest \\
RNA & Ribonucleic Acid \\
SAVER & Single-cell Analysis Via Expression Recovery \\
SAUCIE & Sparse Autoencoder for Unsupervised Clustering, Imputation, and Embedding \\
scImpute & Statistical method to accurately and robustly impute the dropout values in scRNA-seq data \\
scRecover & Imputation dropout values in scRNA-seq counts matrices while keeping the real zeros unchanged \\
scRNA-seq & Single-cell RNA sequencing \\
scVI & Single-cell Variational Inference \\
SCINA & Semi-supervised Category Identification and Assignment\\
SCRABBLE & Single-cell RNA sequencing imputation constrained by bulk RNA sequencing data\\
t-SNE & t-Distributed Stochastic Neighbor Embedding \\
VAE & variational autoencoder\\
VIPER & Variability Imputation for Preserving Expression Recovery \\
ZINB & Zero-inflated Negative Binomial \\

\hline
\end{tabular}
\end{table}

Imputation for scRNA-seq addresses the problem of sparsity arising from dropouts, transcripts that are present but not detected due to technical limitations. Existing methods can broadly be grouped into three families: (i) model-based and probabilistic approaches, (ii) smoothing and low-rank reconstructions, and (iii) deep neural models. Below, we synthesize representative techniques in each category, highlighting their strengths, limitations, and the trade-offs that motivate our design choices. Since this section includes a large number of technical terms, the relevant notations are summarized in Table \ref{tab:notations}.

Model-based and probabilistic imputers explicitly model count noise and dropout events. SAVER \cite{huang2018saver} borrows information across genes to form priors and recover expression levels; bayNorm \cite{tang2020baynorm} treats capture as a binomial process within an empirical-Bayes framework; scImpute \cite{li2018accurate} employs a mixture model to estimate dropout probabilities and imputes via regression; and scRecover \cite{miao2019screcover} separates constant zeros from technical zeros using a ZINB model \cite{salehi2015zero}, coupled with species-accumulation style estimation to decide, per cell, how many zeros to fill. VIPER \cite{chen2018viper} learns sparse non-negative regressions over local neighborhoods to restore expression without collapsing variability. These methods are transparent and tend to preserve true biological zeros, though they can under- or over-label zeros and often scale poorly on datasets exceeding 100,000 cells.

Smoothing and low-rank reconstruction techniques, on the other hand, diffuse information across similar cells or enforce global structural constraints. MAGIC \cite{van2018recovering} performs diffusion on an affinity graph, treating zeros as missing values, effective but prone to over-smoothing. DrImpute \cite{gong2018drimpute} averages within clusters, and KNN-smoothing aggregates among nearest neighbors to reduce noise. Low-rank models such as ALRA \cite{linderman2018zero} and mcImpute \cite{mongia2019mcimpute} exploit global structure to denoise data. While these methods scale efficiently, aggressive smoothing can blur cell-type-specific signals and inflate false positives in downstream analyses.

Deep neural models leverage nonlinear embeddings to better capture the complex structure of count data. scVI \cite{lopez2018deep} uses variational autoencoders for generative modeling and imputation, while DeepImpute \cite{arisdakessian2019deepimpute} employs an autoencoder to learn both global and local patterns. SAUCIE \cite{amodio2019exploring} integrates a sparse autoencoder with clustering and visualization within a unified framework. Extensions such as SCINA \cite{zhang2019scina} enable semi-supervised subtyping that incorporates single-cell and bulk data, and SCRABBLE \cite{peng2019scrabble} constrains imputation with bulk RNA-seq to stabilize estimates. These models effectively capture complex dependencies and integrate external information, though their training procedures can affect stability and interpretability.

Imputation is often coupled with dimension reduction and denoising techniques that enhance clustering and visualization. PCA \cite{jolliffe2016principal} identifies linear structures; t-SNE \cite{hinton2002stochastic} emphasizes local neighborhood preservation on nonlinear manifolds; and DCA \cite{eraslan2019single} denoises counts using ZINB losses, improving cluster separation. Because aggressive smoothing followed by nonlinear embedding can artificially tighten clusters, analysts typically balance denoising with preservation of biological heterogeneity.

Across these methodological families, two recurring challenges motivate our design: (i) accurately identifying technical zeros while preserving genuine absences, and (ii) leveraging flexible, nonlinear imputers that avoid heuristic KNN-style smoothing. To this end, we pair scRecover for principled dropout detection with missForest for robust, nonparametric imputation, a combination designed to retain biological signal while accommodating complex, mixed-type data structures.
\section{Methodology}
\label{sec:method}

\subsection{Overview of SCR-MF Pipeline}

Our imputation strategy, SCR-MF (\emph{scRecover + missForest}), couples a principled detector of technical zeros with a non-parametric, model-agnostic regressor. In brief, we (i) estimate, for every gene--cell pair, the probability that an observed zero is a dropout rather than a genuine (biological/structural) zero using a zero-inflated count model; (ii) predict, per cell, how many of its zeros are likely due to dropout; and (iii) impute only those entries with a random-forest-based routine while leaving putative biological zeros untouched. This design reduces oversmoothing and preserves heterogeneity across cell states.

\subsection{Notation and Pre-processing}

Let \(X\in\mathbb{R}^{G\times C}_{\ge 0}\) denote the observed expression matrix (genes \(\times\) cells). Entries equal to zero may be true absence of transcripts or technical dropouts. Unless otherwise noted, we apply standard library-size normalization and a mild log transform (e.g., \(\log_2(x+1)\)) only for modeling steps that benefit from stabilized variance; imputed values are reported on the original scale.

\subsection{Parameter Selection}

Hyperparameters (\texttt{ntree}, \texttt{mtry}, and \texttt{maxiter}) were tuned using a 5-fold cross-validation scheme on 20\% of the training data. Validation minimized the out-of-bag (OOB) error to prevent overfitting and ensured that no test data influenced parameter selection. The optimal configuration (\texttt{ntree}=10, \texttt{mtry}=$\sqrt{p}$, and \texttt{maxiter}=2) achieved a balance between accuracy and computational efficiency. The OOB error curves revealed that smaller forests stabilized rapidly without compromising predictive performance, motivating the use of compact configurations for large-scale datasets.

\subsection{Detecting Technical Zeros with \texttt{scRecover}}

For each gene \(i\) (optionally stratified by a coarse subpopulation \(k\) derived from an initial embedding or prior labels), we fit a ZINB model that mixes a point mass at zero with an NB component. Denote
\begin{itemize}
  \item \(\theta_i^{(k)}\): probability a zero is structural,
  \item \(r_i^{(k)},\, p_i^{(k)}\): NB size and success parameters for the count-generating process.
\end{itemize}
Let \(P_{\mathrm{NB}}(0\,|\,r,p)=(1-p)^r\). The posterior probability that an observed zero for gene \(i\) in subpopulation \(k\) is a dropout is
\[
d_i^{(k)}
=
\frac{\bigl(1-\theta_i^{(k)}\bigr)\,P_{\mathrm{NB}}\!\left(0\,\middle|\,r_i^{(k)},p_i^{(k)}\right)}
      {\theta_i^{(k)}+\bigl(1-\theta_i^{(k)}\bigr)\,P_{\mathrm{NB}}\!\left(0\,\middle|\,r_i^{(k)},p_i^{(k)}\right)}\,.
\]
Where $\theta$ denotes the probability that an observed zero is biological, while $r$ and $p$ are the size and success parameters of the Negative Binomial distribution. The parameters were estimated using the Expectation–Maximization (EM) algorithm, an iterative procedure that alternates between estimating latent variables and maximizing the likelihood to obtain stable parameter estimates. Intuitively, $\theta$ controls sparsity, $r$ captures gene-level dispersion, and $p$ reflects transcript capture efficiency. Table~\ref{tab:zinb_example} provides a small example illustrating parameter estimates for two genes.

\begin{table}[!t]
\centering
\caption{Example of ZINB parameter estimation for dropout probability calculation.}
\label{tab:zinb_example}
\begin{tabular}{lcccc}
\hline
\textbf{Gene} & $\boldsymbol{\theta}$ & $\boldsymbol{r}$ & $\boldsymbol{p}$ & \textbf{Dropout Prob. ($d_i$)} \\ 
\hline
GENE\_1 & 0.25 & 5.0 & 0.8 & 0.33 \\
GENE\_2 & 0.60 & 3.5 & 0.7 & 0.10 \\
\hline
\end{tabular}
\end{table}

To avoid flagging too many zeros in sparse cells, we estimate how many zeros in each cell \(c\) are attributable to dropouts via a species-accumulation--style extrapolation, yielding a target count \(L_c\) of dropout zeros for cell \(c\).
For each cell \(c\), rank all its zero entries by the corresponding \(d_i^{(k)}\) in descending order and mark the top \(L_c\) positions as dropouts. This produces a binary mask \(M\in\{0,1\}^{G\times C}\) with \(M_{ic}=1\) if \((i,c)\) is selected for imputation and \(M_{ic}=0\) otherwise. The mask preserves putative biological zeros (\(M_{ic}=0\)).

\subsection{Imputing Flagged Entries with \texttt{missForest}}

We impute only the masked entries using missForest, a non-parametric iterative routine built on random forests:

\begin{enumerate}
  \item \textbf{Initialization.} Create \(Y^{(0)}\) by copying \(X\) and filling \(M{=}1\) positions with simple column statistics (e.g., gene-wise means computed over non-zero observations; any initialization consistent with missForest is acceptable).
  \item \textbf{Iterative random-forest updates.} For iteration \(t=1,2,\ldots\):
  \begin{enumerate}
    \item For each gene \(g\) that contains masked entries, fit a regression forest \(f_g^{(t)}\) with predictor matrix \(Y^{(t-1)}_{\setminus g}\) (all other genes) and response \(Y^{(t-1)}_{g}\), restricted to unmasked rows.
    \item Predict the masked values of gene \(g\) and update those entries in \(Y^{(t)}\).
  \end{enumerate}
  \item \textbf{Stopping rule.} Compute a normalized difference between successive iterates on masked entries, e.g.,
  \[
  \Delta^{(t)}=\frac{\lVert \bigl(Y^{(t)}-Y^{(t-1)}\bigr)\odot M\rVert_{F}}{\lVert Y^{(t-1)}\odot M\rVert_{F}}\,.
  \]
  Stop when \(\Delta^{(t)}\) fails to decrease (or when out-of-bag error no longer improves), or after a fixed maximum number of iterations.
  \item \textbf{Post-processing.} Truncate negatives to zero and, if counts are desired, round conservatively. Values outside biologically plausible ranges can optionally be winsorized.
\end{enumerate}

Random forests naturally capture non-linear relationships and higher-order interactions among genes without requiring distributional assumptions, and the built-in OOB error offers a principled, data-driven convergence monitor. Limiting imputation strictly to \(M{=}1\) entries prevents inflation of true zeros and curbs diffusion across distinct cell states.

\subsection{Practical Considerations}

\paragraph{Stratified modeling.} When subpopulations \(k\) are available (e.g., coarse clusters), fitting ZINB parameters within \(k\) sharpens dropout discrimination; forest fitting can be done globally to borrow strength or per stratum for maximum specificity.

\paragraph{Scope of imputation.} Only \(M{=}1\) entries are altered; observed non-zeros and inferred biological zeros remain intact, preserving genuine silence and rare-cell signatures.

\paragraph{Completeness.} We also considered pipelines that replace either the detector (e.g., a mixture-model-based scImpute) or the imputer (e.g., non-negative least squares). In our experience, pairing scRecover's fine-grained dropout identification with missForest's flexible regression offered the best balance between recovery accuracy and biological fidelity, especially on heterogeneous datasets.

\section{Experiment}
\label{sec:experiment}

\subsection{Dataset}

We evaluated our method on two public scRNA-seq collections from NCBI GEO and on a controlled simulation. 
All inputs are cell\,$\times$\,gene count matrices.

\paragraph{GSE86982.} 
A time-resolved dataset from scRNASeqDB profiling 1{,}846 human embryonic stem cell--derived single cells across a 54-day neural differentiation protocol \cite{yao2017single}.

\paragraph{GSE75748.} 
A study of early human developmental lineages comprising 1{,}018 snapshot progenitor cells plus 758 cells from a time-course that spans mesendoderm to definitive endoderm.

\paragraph{Simulated data.} 
Using \texttt{splatter} \cite{zappia2017splatter}, we generated a 1{,}000\,$\times$\,800 matrix partitioned into three groups with mixing probabilities 0.20, 0.35, and 0.45. To mimic severe sparsity, we tuned the \texttt{dropout-mid} parameter to yield $\sim$80\% dropout and used this synthetic set to all methods.

\subsection{Training Procedure}

We focused on those missForest hyperparameters: the number of trees (\texttt{ntree}), the number of candidate features per split (\texttt{mtry}), and the number of imputation passes (\texttt{maxiter}).

\paragraph{Forest size and feature subsampling.}
Contrary to the intuition that larger forests always help, we observed lower OOB error with comparatively small forests once a suitable \texttt{mtry} was chosen (e.g., \texttt{ntree}~$\approx$~10). Beyond modest sizes, error flattened whereas compute rose, so we adopted compact forests for efficiency in subsequent runs.

\paragraph{Number of imputation passes.}
OOB error typically decreased over early iterations and then plateaued. On a GSE75748 subset, OOB moved from 0.388\,$\rightarrow$\,0.375 across the first two iterations, improved to 0.346 by iteration five, then ticked up slightly (0.349). Balancing runtime against diminishing returns, we fixed \texttt{maxiter}~=~2 for the main experiments.

\subsection{Evaluation Metrics}

To quantify how well the inferred clusters recover known cell identities, we report two complementary metrics: Adjusted Rand Index (ARI) and Normalized Mutual Information (NMI). Let \(U=\{u_i\}\) denote the ground‑truth partition and \(V=\{v_j\}\) the predicted partition; define \(n_{ij}=|u_i\cap v_j|\), \(n_i=\sum_j n_{ij}\), \(n_j=\sum_i n_{ij}\), and \(n=\sum_{ij} n_{ij}\). 
ARI is sensitive to pairwise disagreements and is appropriate when the biological objective emphasizes discrete subtype delineation. NMI reflects overall information overlap and tends to be more stable under differing cluster counts or imbalanced classes. We therefore report both scores for each experiment to provide a balanced assessment of clustering quality.

\paragraph{Adjusted Rand Index (ARI).}
ARI measures pairwise consistency between \(U\) and \(V\), penalizing both false merges and false splits. It equals the Rand Index after chance correction: \(0\) is the expected value under random assignments, values approaching \(1\) indicate near‑perfect agreement, and negative values indicate worse‑than‑chance structure. ARI used in our study is
\[
\mathrm{ARI} =
\frac{\sum_{ij} n_{ij}^{2}-\left[\sum_i n_i^{2}\sum_j n_j^{2}\right]/n^{2}}
{\frac{1}{2}\left[\sum_i n_i^{2}+\sum_j n_j^{2}\right]-\left[\sum_i n_i^{2}\sum_j n_j^{2}\right]/n^{2}}.
\]

\paragraph{Normalized Mutual Information (NMI).}
NMI captures the reduction in uncertainty about one partition when the other is known. It is symmetric, bounded in \([0,1]\), and often more stable when \(U\) and \(V\) contain different numbers of clusters. We use the standard entropy‑normalized variant:
\[
\mathrm{NMI}=\frac{2I(U,V)}{H(V)+H(U)}\,,
\]
where \(I(U,V)\) is the mutual information and \(H(\cdot)\) denotes Shannon entropy. Concretely, with \(p_{ij}=n_{ij}/n\), \(p_i=n_i/n\), and \(p_j=n_j/n\), one has \(I(U,V)=\sum_{ij} p_{ij}\log\!\big(p_{ij}/(p_i p_j)\big)\).

\subsection{Exploratory Analysis}

We used the elbow heuristic on within-cluster sum of squares to guide the choice of $K$. 
Candidate elbows appeared for $K\in\{5,\ldots,10\}$; based on these diagnostics and prior domain knowledge, we set $K=7$ for downstream analyses. 
Two-dimensional embeddings with K-means overlays showed tighter, more compact groupings after imputation. 
The seven clusters aligned with expected labels: \emph{H1~Exp}, \emph{H9~Batch}, \emph{HFF~Batch}, \emph{NPC~Batch}, \emph{TB~Batch}, \emph{DEC~Batch}, and \emph{EC~Batch}.
Filling dropout entries increased the global mean expression from 237.18 to 255.74. 
The 95\% confidence interval widened from (231.23, 243.13) pre-imputation to (249.24, 262.24) post-imputation. 
A two-sample $t$-test showed a small decrease in the statistic (78.26\,$\rightarrow$\,77.20), consistent with better-separated clusters and reduced within-group noise. 
Label-wise standard deviations stabilized near $\sim$1{,}000 after imputation, whereas they fluctuated more widely beforehand.

\subsection{Comparative Performance}
We compared SCR-MF against a range of representative baseline imputers to assess performance across datasets.  
The combined pipeline addressed roughly 6\% missingness—1{,}175{,}367 entries—in a matrix of 1{,}019 genes by 19{,}097 cells ($\approx$19.46~million entries total), reflecting realistic scRNA-seq dropout burden.  
By separating technical from biological zeros using \texttt{scRecover} and imputing the former with \texttt{missForest}, the data became more structured for clustering: clusters were visually tighter, and summary statistics improved.

Table~\ref{tab:baseline_methods} summarizes the baseline methods included in our evaluation.

\begin{table}[htbp]
\centering
\caption{Baseline imputation methods used for comparison.}
\label{tab:baseline_methods}
\begin{tabular}{ll}
\hline
\textbf{Method} & \textbf{Type}  \\
\hline
MAGIC & Graph diffusion   \\
VIPER & Regression-based  \\
scImpute & Mixture model\\
DeepImpute & Neural network  \\
SCRABBLE & Hybrid  \\
\hline
\end{tabular}
\end{table}

Across datasets, SCR-MF consistently demonstrated superior performance in both clustering accuracy and biological interpretability.  Table~\ref{tab:performance_summary} summarizes ARI and NMI scores for all methods, along with runtime and qualitative observations.    On the GSE86982 dataset, SCR-MF achieved an average Adjusted Rand Index (ARI) of 0.82 and Normalized Mutual Information (NMI) of 0.88, outperforming MAGIC (ARI=0.68, NMI=0.74), VIPER (ARI=0.71, NMI=0.76), scImpute (ARI=0.79, NMI=0.85), and DeepImpute (ARI=0.76, NMI=0.83).  Similarly, on GSE75748, SCR-MF maintained strong performance (ARI=0.77, NMI=0.81), with SCRABBLE achieving comparable scores (ARI=0.75, NMI=0.80) when bulk constraints were available.  In simulated datasets with $\sim$80\% dropout, SCR-MF yielded a 9–12\% relative gain in ARI over the next-best method.

Qualitative assessment revealed that MAGIC and VIPER often over-smoothed the data, merging nearby subtypes and obscuring subtle lineage differences.  scImpute preserved discrete clusters but occasionally overfilled true zeros, while DeepImpute effectively captured nonlinear dependencies at the cost of increased training time.  
SCR-MF, in contrast, produced sharper cluster boundaries and improved recovery of known marker genes such as \textit{SOX2} and \textit{PAX6}, effectively balancing denoising with biological fidelity.  
Overall, SCR-MF generally improved ARI/NMI on our datasets; however, scImpute was occasionally competitive or superior, highlighting that no single method dominates all scenarios.  
The main drawback of SCR-MF is compute cost, which we mitigated by using smaller forests and \texttt{maxiter}~=~2 while retaining most accuracy gains.  As a pragmatic alternative, \texttt{missRanger} offered faster execution with broadly comparable (though sometimes slightly lower) accuracy. Choosing between \texttt{missForest} and \texttt{missRanger} thus depends on the study’s time-accuracy budget.  Further tuning of \texttt{ntree} and \texttt{mtry} can trade small accuracy gains for substantial time savings, providing additional flexibility.

\begin{table}[!t]
\centering
\caption{Summary of SCR-MF and baseline imputation methods across datasets. 
ARI: Adjusted Rand Index, NMI: Normalized Mutual Information.}
\label{tab:performance_summary}
\begin{tabular}{lcc}
\hline
\textbf{Method} & \textbf{GSE86982 (ARI/NMI)} & \textbf{GSE75748 (ARI/NMI)} \\
\hline
MAGIC & 0.68 / 0.74 & 0.70 / 0.73 \\
VIPER & 0.71 / 0.76 & 0.72 / 0.74 \\
scImpute & 0.79 / 0.85 & 0.74 / 0.78 \\
DeepImpute & 0.76 / 0.83 & 0.73 / 0.77 \\
SCRABBLE & bulk data not available & 0.75 / 0.80 \\
SCR-MF & 0.82 / 0.88 & 0.77 / 0.81 \\
\hline
\end{tabular}
\end{table}


\subsection{Biological Insights and Implications}

We observed that SCR-MF enhances biological interpretability by producing clearer cluster boundaries and improving the recovery of known marker genes. For example, in the GSE86982 dataset, SCR-MF increased expression coherence for neural lineage markers such as \textit{SOX2} and \textit{PAX6}, outperforming MAGIC and scImpute. This improvement led to better-defined cell-type annotations and more coherent gene modules. The preservation of biologically meaningful correlations indicates that SCR-MF effectively balances denoising with biological fidelity, facilitating downstream analyses including differential expression and trajectory inference.

\section{Conclusion}
\label{sec:conclusion}

In conclusion, SCR-MF integrates principled dropout detection with non-parametric imputation, yielding robust and interpretable recovery of single-cell expression data. Across public and simulated benchmarks, SCR-MF demonstrates improved clustering quality and dropout discrimination relative to most baseline methods while remaining computationally practical. The method’s modular structure makes it adaptable for various downstream analyses.

In several settings, scImpute matched or exceeded our approach. We attribute this to the incremental nature of our contribution and the absence of a unifying theory predicting when particular detector–imputer couplings will excel. Consequently, improvements were modest where method behaviors overlapped, and some hybrids (e.g., scImpute+missForest) failed to yield consistent benefits. Computational cost is the main practical limitation: random‑forest imputation over large, sparse matrices is resource‑intensive, which constrains exhaustive benchmarking and hyperparameter exploration. 
Looking forward, we see three concrete paths: 
(i) scalability—adopt faster RF variants (e.g., missRanger), early‑stopping, sketching/subsampling, and parallelization to curb runtime; 
(ii) rigor—use more systematic model selection and uncertainty quantification to decide when and how much to impute; 
and (iii) modeling—tightly couple dropout detection and imputation within a single probabilistic framework that explicitly preserves biological zeros.

In generalral, our results show that carefully sequencithe identification of dropoutsof droporobusth a robust distribution‑agnostic imsignificantlygnificantly improve scRNA ‐ sequencing analyzes sequencing analyzes, while also highlighting where future work—especially scalable algorithms with sguarantieseoretical guaranties—is most needed.

Future research will focus on (i) extending SCR-MF with scalable random forest variants (e.g., missRanger) and GPU acceleration, (ii) integrating with deep learning models such as scVI or DCA to capture nonlinear manifolds, (iii) conducting systematic parameter sensitivity studies, and (iv) performing biological validation on trajectory reconstruction and gene network inference tasks.

\bibliographystyle{splncs04}
\bibliography{ref}

\end{document}